\documentclass[conference]{IEEEtran}
\IEEEoverridecommandlockouts

\usepackage[utf8]{inputenc}
\usepackage{textcomp}
\usepackage[style=ieee,isbn=false,url=false,doi=false]{biblatex}
\usepackage{amsmath,amssymb,amsfonts}
\usepackage{siunitx}
\usepackage{array}
\usepackage{graphicx}
\usepackage{stfloats}
\usepackage{caption}
\usepackage[labelformat=simple]{subcaption}

\usepackage{placeins}
\usepackage[dvipsnames]{xcolor}
\usepackage[inline]{enumitem}
\usepackage{booktabs}
\usepackage{multirow}
\usepackage{algorithmic}
\usepackage{algorithm}

\usepackage[accsupp]{axessibility}  

\usepackage[bookmarks=false]{hyperref}

\usepackage[capitalize]{cleveref}
\Crefname{figure}{Fig.}{Figs.}

\def\BibTeX{{\rm B\kern-.05em{\sc i\kern-.025em b}\kern-.08em
    T\kern-.1667em\lower.7ex\hbox{E}\kern-.125emX}}

\begin{document}

\title{{LOD-Net}: Locality-Aware {3D} Object Detection Using Multi-Scale Transformer Network} 

\author{%
\IEEEauthorblockN{%
Mustaqeem Khan\IEEEauthorrefmark{1}, 
Aidana Nurakhmetova\IEEEauthorrefmark{2},
Wail Gueaieb\IEEEauthorrefmark{3} and Abdulmotaleb Elsaddik\IEEEauthorrefmark{3}}
\IEEEauthorblockA{\IEEEauthorrefmark{1}College of Information Technology-United Arab Emirates University (UAEU), Al Ain, UAE}
%
\IEEEauthorblockA{\IEEEauthorrefmark{2}Mohamed bin Zayed University of Artificial Intelligence, Abu Dhabi, UAE}
%
\IEEEauthorblockA{\IEEEauthorrefmark{3}School of Elec. Eng. \& Comp. Sc., University of Ottawa, Ottawa, ON, Canada\\
Email: Mustaqeemkhan@uaeu.ac.ae, Aidana.Nurakhmetova@mbzuai.ac.ae, \{wgueaieb, elsaddik\}@uottawa.ca}%
}

\maketitle

\begin{abstract}
3D object detection in point cloud data remains a challenging task due to the sparsity and lack of global structure inherent in the input. In this work, we propose a novel Multi-Scale Attention (MSA) mechanism integrated into the 3DETR architecture to better capture both local geometry and global context. Our method introduces an upsampling operation that generates high-resolution feature maps, enabling the network to better detect smaller and semantically related objects. Experiments conducted on the ScanNetv2 dataset demonstrate that our 3DETR + MSA model improves detection performance, achieving a gain of almost 1\% in mAP@25 and 4.78\% in mAP@50 over the baseline. While applying MSA to the 3DETR-m variant shows limited improvement, our analysis reveals the importance of adapting the upsampling strategy for lightweight models. These results highlight the effectiveness of combining hierarchical feature extraction with attention mechanisms in enhancing 3D scene understanding.
%
\end{abstract}


\section{Introduction}
3D object detection presents significant challenges due to the unstructured and sparse nature of point cloud data. A point cloud consists of discrete points sampled from surfaces within a 3D environment, often resulting in an irregular and unordered representation of the scene. Despite these challenges, 3D object detection has seen widespread application across various domains, including robotics~\cite{3D_applic_robotics,3D_applic_robotics2}, healthcare~\cite{3D_applic_healthcare,3D_applic_healthcare2, AbdullahAlAnziAlSharhan2018}, autonomous driving~\cite{3D_applic_driving}, and augmented and virtual reality~\cite{3D_applic_AR}.

\begin{figure}[htb]
\centering
\includegraphics[width=\columnwidth]{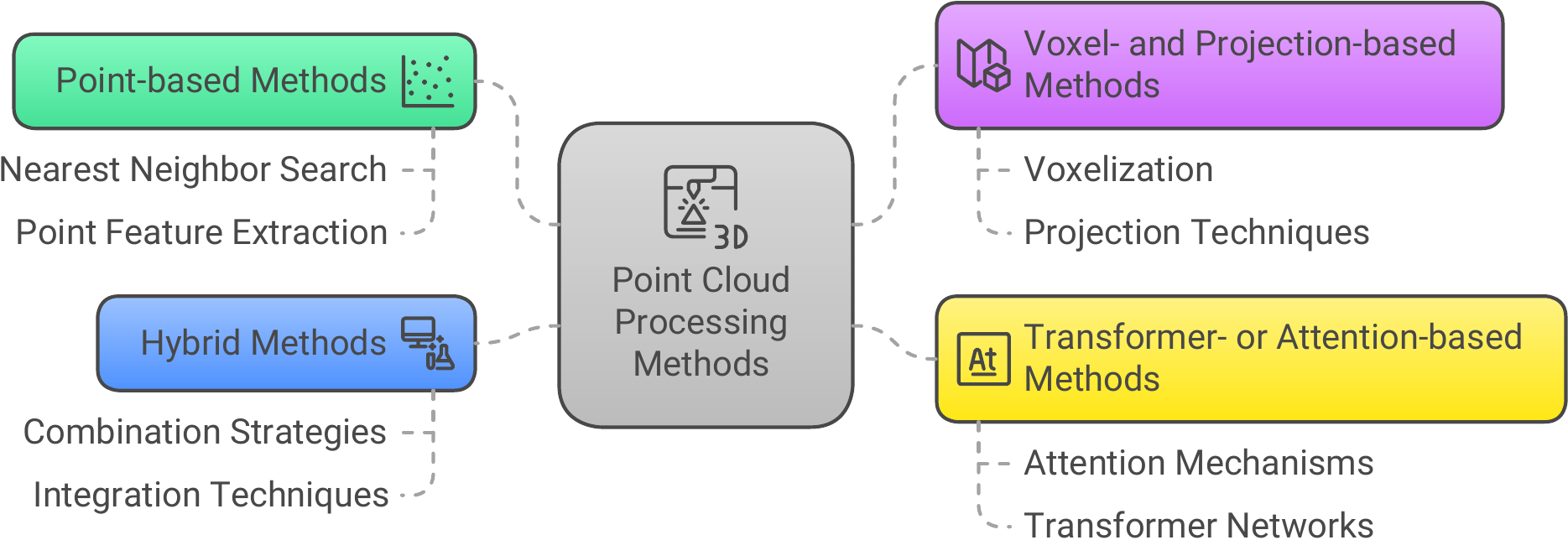}
\caption{3D point cloud processing methods}
\label{fig:3DPC-processing-methods}
\end{figure}

In this work, we focus specifically on indoor 3D object detection, distinguishing our setting from outdoor applications, such as those encountered in self-driving vehicle systems.
Numerous deep learning methodologies have been proposed to effectively extract features from 3D point clouds. Among them, PointNet++~\cite{PointNet2} has demonstrated strong capabilities in hierarchical feature learning from irregular 3D data. Unlike conventional deep learning inputs, such as structured 2D images, sequential text, and numerical data, 3D point clouds require specialized processing techniques to account for their inherent irregularity and sparsity.
As summarized in \Cref{fig:3DPC-processing-methods}, existing approaches to 3D point cloud processing can be broadly categorized into the following groups
\begin{enumerate*}[before=\unskip{: }, itemjoin={{; }}, itemjoin*={{, and }}]
    \item Point-based methods
    \item Voxel- and projection-based methods
    \item Transformer- or attention-based methods
    \item Hybrid methods
\end{enumerate*} 
In Section~\ref{related_work}, we review representative works from each of these categories, providing context for our proposed approach.

Deep learning has driven significant advancements in 3D scene understanding over the past several years. Researchers worldwide have proposed novel architectures and techniques for tasks such as shape analysis, semantic segmentation, object alignment, grasp detection, and registration. Notable studies, including~\cite{DGCNN_2018,li2018pointcnn,jiang2018pointsift}, have explored network designs and convolutional operations specifically tailored for 3D point cloud data. Other works, such as~\cite{zeng20173dmatch} and~\cite{avetisyan2018scan2cad}, have leveraged RGB-D reconstructions and CAD model alignments to enhance scene understanding.

Several survey papers~\cite{Xiao_survey, NiceSurvey_2020, xiao2023survey} provide comprehensive overviews of deep learning methods for 3D point cloud processing. In recent years, self-attention mechanisms and Transformer-based architectures have been increasingly adopted for 3D object recognition and segmentation tasks. Traditionally, the irregular and complex nature of 3D scenes necessitated additional data manipulations, such as projection or voxelization, which introduced computational overhead and complexity. However, the ability of Transformers to efficiently capture long-range dependencies has enabled direct processing of raw point cloud inputs, eliminating the need for such preprocessing.

Transformer architectures offer a global contextual view of a scene, enabling networks to effectively model relationships between different object clusters. In 3DETR~\cite{3DETR}, the first end-to-end 3D object detection framework based on Transformers, the encoder is designed to learn point-to-point relations, while the decoder focuses on modeling point-to-object and object-to-object associations. However, due to the extensive complexity of the scene and computational constraints, the original number of points fed into the model is downsampled by nearly 20 times. This aggressive downsampling results in substantial information loss, making it challenging even for powerful architectures to accurately capture the shapes and attributes of scene objects.

To address this limitation, we propose a technique that enhances feature representation by introducing an upsampling operation that generates high-resolution feature maps. These upsampled features are subsequently processed by the multi-head attention mechanism in the decoder, alongside the original feature points, allowing the network to preserve fine-grained details crucial for object detection more effectively.
The primary contributions of this work are as follows:
\begin{itemize} 
    \item We introduce a novel locality-aware 3D object detection architecture that incorporates a Multi-Scale Attention (MSA) mechanism, designed to extract fine-grained features, particularly benefiting smaller object categories efficiently. 
    \item We develop a lightweight and enhanced version of the 3DETR decoder, leading to improvements in detection performance on the 3DETR (in terms of mAP@25 and mAP@50) and 3DETR-m (in terms of mAP@25) models, with notable gains in detecting small-scale objects (as detailed in \Cref{sec:experiments}). 
\end{itemize}

\section{Related Work}
\label{related_work}
\subsection{Point-Based Approaches}
PointMixer~\cite{PointMixer_2021} introduces a versatile operator for point sets, enabling the extraction of shared structural information across 3D scenes. In this framework, the token-mixing multi-layer perceptrons (MLPs) commonly used in similar architectures are replaced with a softmax-based mixing operation, enhancing feature aggregation.
EdgeConv~\cite{DGCNN_2018} adopts a graph-based convolutional approach, where each network layer processes local neighborhoods within the point cloud, allowing the network to learn local geometric structures dynamically.
The 3D-MPA method~\cite{3D-MPA2020} leverages graph convolutional networks to model inter-proposal relationships, thereby improving the network's ability to reason about object interactions and per-point feature representations.
BRNet~\cite{BRNet2021} is inspired by the back-tracing mechanism of the traditional Hough voting algorithm. It refines the estimation of representative points around voted centers and re-evaluates seed points to achieve more accurate object localization.

\subsection{Projection/Voxel-Based Approaches}
MV3D~\cite{MV3D} integrates LIDAR point clouds and RGB images through a combination of 2D and 3D convolutional operations to extract rich features and predict oriented 3D bounding boxes. The network achieves accurate object detection by employing a proposal generation subnetwork that projects the 3D point cloud into a bird’s-eye view representation.
FCAF3D~\cite{FCAF3D} introduces a fully convolutional approach for 3D object detection that eliminates the reliance on predefined anchor boxes. Instead, the network directly regresses bounding box parameters from the point cloud data, enabling precise and efficient object detection.
GSDN~\cite{GSDN} presents a novel detection framework that combines sparse detection networks with generative adversarial networks (GANs). By leveraging GANs to generate high-resolution object proposals, the method significantly improves detection performance, particularly in scenarios with limited training data.

\subsection{Transformer-Based Approaches}
The Point Cloud Transformer (PCT)~\cite{PCT_2021} is specifically designed for point cloud analysis, leveraging self-attention mechanisms and positional encoding to capture long-range dependencies while preserving spatial information effectively.
Point Transformer~\cite{Point_Trans} targets classification and dense prediction tasks, serving as a potential backbone for a wide range of 3D scene understanding models. A key innovation of this architecture is the use of vector self-attention, in contrast to traditional scalar attention, where relational information is modeled through a subtraction-based function.
Pointformer~\cite{Pointformer2020} introduces a transformer-based architecture that adopts a U-Net-like structure, utilizing the global modeling capabilities of transformers to capture long-range dependencies effectively for 3D scene analysis.

\subsection{Hybrid Architectures}
In 3D point cloud processing, hybrid architectures have emerged to combine the strengths of multiple paradigms. These architectures typically integrate voxel-based methods with attention mechanisms or merge point-based and graph-based approaches with voxelization and attention modules, resulting in unified and robust networks.
An example of this trend is VoxSet~\cite{VoxSetTr}, where 3D voxels are processed using a Transformer model to capture both local and global contextual information. The architecture introduces a Voxel-based Set Attention (VSA) module, which replaces standard self-attention with two cross-attention mechanisms applied within each voxel, enhancing feature aggregation.
The Point-Voxel Transformer (PVT)~\cite{PVT_2021} also follows a hybrid strategy. Initially, the point cloud is voxelized, and local features are extracted using 3D convolutional neural networks. Rather than relying solely on voxel features, PVT further refines the representation through a dedicated Point-Voxel Transformer, composed of multiple Transformer layers.
PatchFormer~\cite{Zhang_2022_CVPR} presents an efficient adaptation of the Point Transformer model by introducing a Patch Attention mechanism. This approach partitions the point cloud into patches, enabling more efficient computation while preserving critical spatial relationships. By focusing on patch-level rather than point-level interactions, PatchFormer achieves competitive performance in tasks such as object classification and part segmentation, while significantly reducing computational complexity.
Group-Free 3D~\cite{Group-free-3d} proposes a fully point-based Transformer network that operates directly on raw point clouds, eliminating the need for grouping or voxelization. This design enables the model to capture fine-grained relationships and dependencies between individual points, simplifying the pipeline while maintaining strong detection performance.

\section{Proposed Approach}
Our proposed system design is based on the seminal work of \citeauthor{Transformer} in \cite{Transformer}, which revolutionized feature extraction and representation learning by replacing traditional convolutional operations with attention mechanisms. This paradigm shift has had a profound impact on 2D object detection, where transformer-based architectures have demonstrated strong performance in accurately predicting bounding boxes. Transformers excel at modeling object relations and capturing global context, eliminating the need for manual anchor design and non-maximum suppression post-processing steps.

The success of 2D Vision Transformers, which process image patch sets, has spurred the development of transformer-based 3D object detectors that can capture long-range relations among set elements. 
By employing encoder-decoder architectures with multi-head attention, skip connections, layer normalization, and MLPs, these detectors can eliminate the need for additional proposal sampling or grouping procedures. 
Self-attention, a key component of transformers, involves scaling dot products between key, query, and value matrices to capture long-range relations and global context. 

Unlike convolution filters that operate on fixed pixel positions, transformers lack positional information as they process sets without a specific order. To address this, positional encoding is introduced, where learnable parameters representing positions are added to the features after encoding through cosine and sine functions. Different works employ various methods for positional encodings, such as Fourier positional encoding in 3DETR~\cite{3DETR}, linear projection via 1D convolution or MLP in Group-free 3D~\cite{Group-free-3d}, and passing relative position encoding to MLP in Point Transformer~\cite{Point_Trans}.
 
\subsection{Multi-scale Transformer Network}
\begin{figure}[htb]
\centering
\includegraphics[width=\columnwidth]{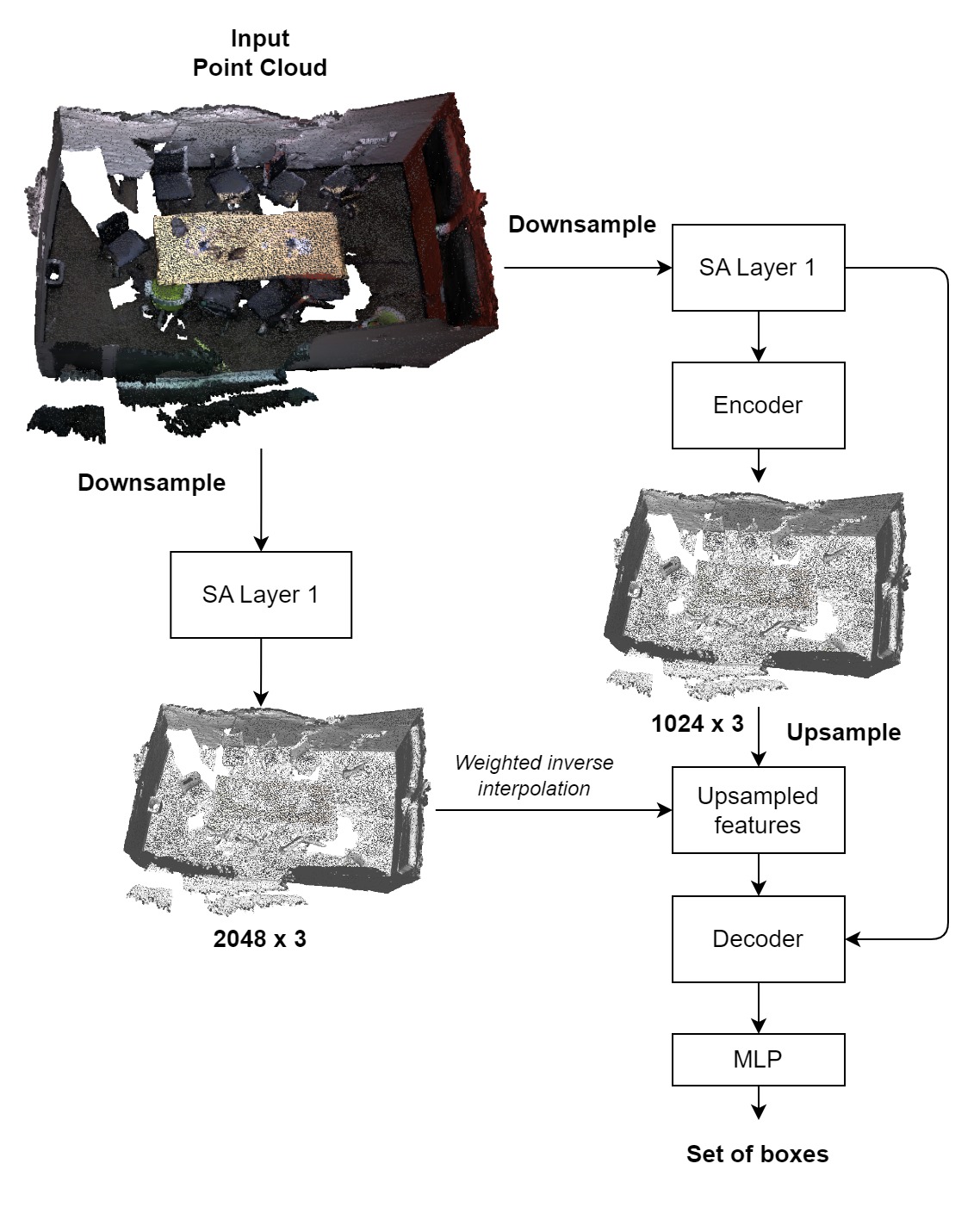}
\caption{The architecture of 3DETR model with multi-scale feature extraction}
\label{fig:main_design}
\end{figure}

The proposed system architecture is illustrated in \Cref{fig:main_design}. The input point cloud is first processed by a set abstraction (SA) layer from PointNet++, where the original 40,000 points are downsampled to a significantly lower number, typically 1,024 or 2,048 points. These sampled points, selected via farthest point sampling (FPS), serve as a compact representation of the entire scene and are subsequently fed into the transformer encoder layers. In parallel, a denser subset of points is sampled using a second SA layer to support the upsampling operation. The transformer decoder then processes the resulting upsampled features to refine the object predictions.

\subsection{Feature Upsampling}
\begin{figure}[htb]
\centering
\includegraphics[width=\columnwidth]{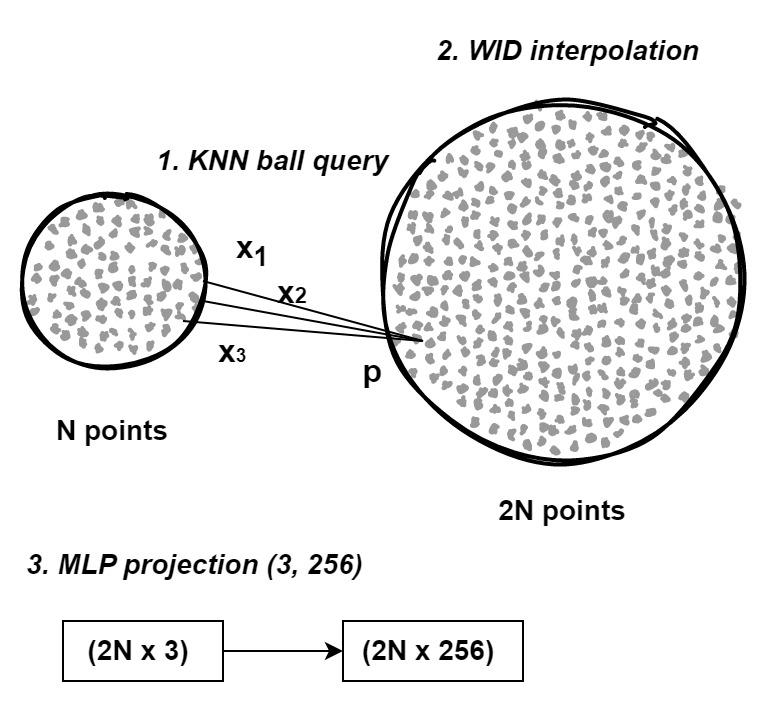}
\caption{Upsampling procedure in the proposed multi-scale transformer-based model.}
\label{fig:upsampler1}
\end{figure}

As illustrated in \Cref{fig:upsampler1}, the upsampling algorithm consists of three main steps. First, for each sampled point \textbf{p}, the three nearest neighbors $(x1, x2, x3)$ are identified from a set of $N$ representing the number of encoder points, using the k-nearest neighbors (k-NN) algorithm based on Euclidean distance. These sampled points are drawn from a larger set of $2N$ points, as depicted in the figure (note that the relative sizes of the spheres are for illustrative purposes only; in practice, the point cloud distribution is sparse and non-uniform). Second, weighted inverse distance (WID) interpolation is performed between each query point and its neighbors to produce the upsampled features. Finally, a multi-layer perceptron (MLP) projects the resulting features into a higher-dimensional space, mapping the 3D input channels to a 256-dimensional feature vector that is subsequently fed into the decoder. The upsampling procedure is presented more formally in \Cref{alg:cap}.

\begin{algorithm}
\caption{Upsampling implementation}\label{alg:cap}
\begin{algorithmic}
\ENSURE $\text{xyz, features} = \text{point\_clouds}$
\STATE $\text{pre\_enc\_xyz} \gets \text{pre\_encoder(xyz, features)}$
\REQUIRE $\text{sampled\_points} = \text{pre\_enc\_xyz}$
\WHILE{$k \leq \text{sampled\_points.length}$}
    \STATE data $\gets$ point\_features[k]
    \STATE neighbors $\gets$ NearestNeighbors (3, `ball\_tree')
    \STATE query\_points $\gets$ sampled\_points[k]
    \STATE indices $\gets$ nbrs.kneighbors (query\_points, 3)
    \STATE neighbors $\gets$ data[indices]
    \STATE wid $\gets$ WID (neighbors, query\_points, 2)
    \STATE upsampled\_feats[k] $\gets$ wid
\ENDWHILE
\STATE feats $\gets$ mlp\_proj(upsampled\_feats)
\end{algorithmic}
\end{algorithm}

\begin{algorithm}
\centering
\caption{MSA implementation} \label{alg:cap2}
\begin{algorithmic}
\STATE WID (neighbors, queries, power):
\STATE dist1 $\gets$ linalg.norm(queries[:] - neighbors[0, :])
\STATE dist2 $\gets$ linalg.norm(queries[:]  - neighbors[1, :])
\STATE dist3 $\gets$ linalg.norm(queries[:]  - neighbors[2, :])
\ENSURE $\epsilon = 1e-8$
\REQUIRE indices $=$ where(dist1 $==$ 0)[0]
\REQUIRE indices $=$ where(dist2 $==$ 0)[0]
\REQUIRE indices $=$ where(dist3 $==$ 0)[0]
\IF {$min(\text{dist1}) \leq \epsilon$} 
  \STATE $min_v \gets$ neighbors[0, indices, 0]
  \STATE dist1[indices, 0] $\gets min_v$
\ENDIF
\IF {$min(\text{dist2}) \leq \epsilon$}
    \STATE $min_v \gets$ neighbors[0, indices, 0]
    \STATE dist2[indices, 0] $\gets min_v$
\ENDIF
\IF {$min(\text{dist3}) \leq \epsilon$}
    \STATE $min_v \gets$ neighbors[0, indices, 0]
    \STATE dist3[indices, 0] $\gets min_v$
\ENDIF 
\STATE $w1 \gets 1 / (\text{dist1}^{\text{power}})$
\STATE $w2 \gets 1 / (\text{dist2}^{\text{power}})$
\STATE $w3 \gets 1 / (\text{dist3}^{\text{power}})$
\STATE $\text{w\_sum} \gets w1 + w2 + w3$
\STATE $\text{w1\_norm} \gets w1 / \text{w\_sum}$
\STATE $\text{w2\_norm} \gets w2 / \text{w\_sum}$
\STATE $\text{w3\_norm} \gets w3 / \text{w\_sum}$
\STATE value $\gets$ w1\_norm * neighbors[0, :] 
+ w2\_norm * neighbors[1, :] + w3\_norm * neighbors[2, :]
\end{algorithmic}
\end{algorithm}

\subsection{MSA with Double Attention Maps}
The self-attention mechanism is a core component of transformer architectures, widely used in natural language processing and computer vision tasks. It enables the model to dynamically weigh the importance of different elements within a sequence or set by computing attention scores. Specifically, each input element, referred to as a \textit{Query}, interacts with all other elements, represented as \textit{Key}-\textit{Value} pairs. The similarity between a Query and each Key is measured through a dot product, which is then scaled and passed through a softmax function to produce normalized attention weights. These weights determine the contribution of each value to the final output, enabling the model to capture complex dependencies and contextual relationships within the input data, as depicted in \Cref{alg:cap2}.

The self-attention operation, originally introduced in~\cite{Transformer} is defined by
\begin{equation}
    Z(Q, K, V) = \sigma\left( \frac{Q \, W^q \, (K \, W^k)^T}{\sqrt{d}} \right) \left( V \, W^v \right) , \label{eq:self-att_dot_product}
\end{equation}
where the matrices $W^q$, $W^k$, and $W^v$, represent the learnable variables, and the softmax operation is denoted by $\sigma$.
The equation computes the dot product between a \textit{Query} and a \textit{Key}, which is subsequently scaled by the square root of the embedding dimension $d$. This scaled attention score, given by $Q K^T / \sqrt{d}$, is then multiplied by a \textit{Value} matrix to produce a weighted sum of the input features. The input matrices for the \textit{Key} ($K$), \textit{Query} ($Q$), and \textit{Value} ($V$) are all derived from the same input $X \in \mathbb{R}^{N\times{d}}$, where $K=Q=V$. 
The final attention map is obtained by applying a formulation similar to~\eqref{eq:self-att_dot_product} across multiple attention heads. The Multi-Head Attention (MHA) mechanism, defined in~\eqref{eq:mhsa}, partitions the input feature set into several heads, with each head independently learning different representations of the input. The outputs from all heads are then concatenated and passed through a linear projection layer to restore the original embedding dimensionality.
\begin{equation}
	\text{MHA} = Z(QW^q_i, KW^k_i, VW^v_i), ~~ \forall i = 1, ... , h  \label{eq:mhsa}
\end{equation}

\begin{figure}[htb]
\centering
\includegraphics[width=0.9\columnwidth]{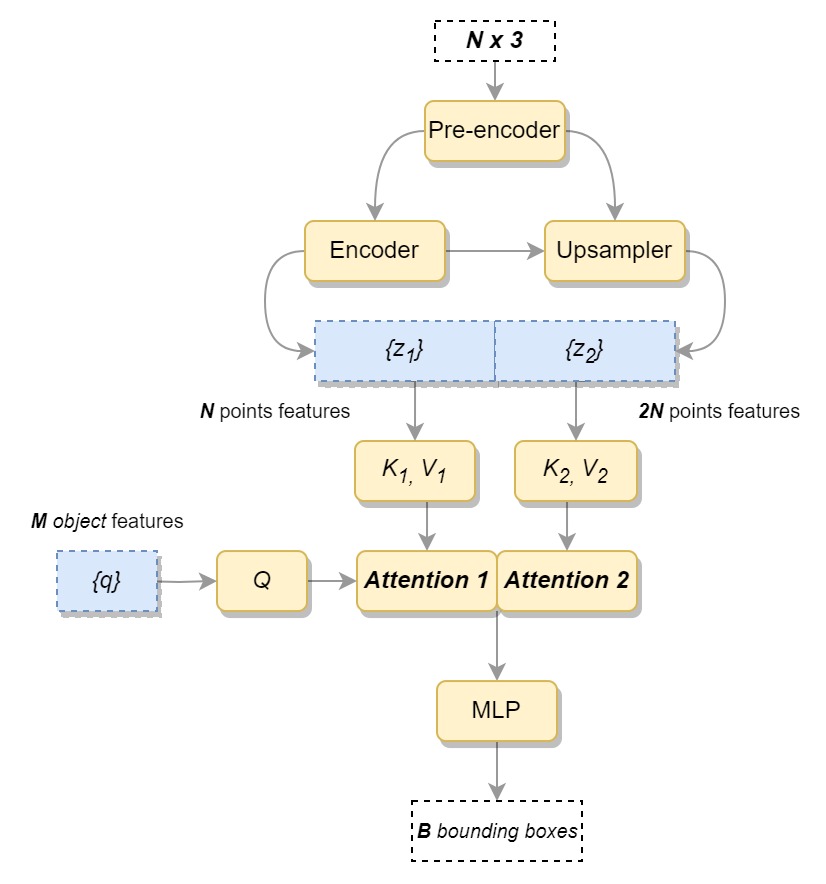}
\caption{Proposed MHA layer architecture.}
\label{fig:upsampler}
\end{figure}

The Multi-Head Attention (MHA) mechanism proposed in this work is illustrated in \Cref{fig:upsampler}.\footnote{It's important to note that the pre-encoder layer preceding this MHA remains the same as the previously described Self-Attention (SA) in layer~1.} It is based on two distinct attention maps, each associated with separate \textit{Key-Value} pairs. In the first group, $(K_1, V_1)$ represents the original point features extracted from the encoder, while $(K_2, V_2)$ corresponds to the upsampled point features. A shared \textit{Query} is used for both groups. Subsequently, two attention maps are generated according to the following expressions:
\begin{align*}
    A_1 = Z(Q, K_1, V_1) = \sigma\left( \frac{Q \, W_1^q \left( K_1 \, W_1^k \right)^T}{\sqrt{d}} \right) \left( V_1 \, W_1^v \right) \\
    A_2 = Z(Q, K_2, V_2) = \sigma\left( \frac{Q \, W_2^q \left( K_2 \, W_2^k \right)^T}{\sqrt{d}} \right) \left( V_2 \, W_2^v \right) 
\end{align*}
The final attention map is obtained by concatenating $A_1$ and $A_2$ along the channel dimension. It is important to note that the weight matrix is logically partitioned across the two Key-Value pairs, effectively mimicking the behavior of two independent attention heads.

\begin{figure}[htb]
\centering
\includegraphics[width=0.3\textwidth]{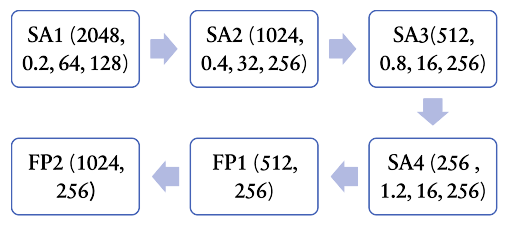}
\caption{PointNet++ structure}
\label{fig:point2}
\end{figure}

\begin{table*}[htb]
\centering
\caption{Per-category evaluation of the proposed architecture applied to the ScanNetV2 dataset at AP@25 IoU.} 
\label{tab:percat2}
\resizebox{\textwidth}{!}{%
\begin{tabular}{|l|cccccccccccccccccc|c|}
\toprule
Method & cab & bed & chair & sofa & table & door & window & bkshf & pic & contr & desk & curtain & fridge & shower & toilet & sink & bath & grbin & mAP \\
\midrule
3DETR (paper) & 50.2 &87.0 &86.0 &87.1 &61.6 &46.6 &40.1 &54.5 &9.1 &62.8 &69.5 &48.4 &50.9 &68.4 &97.9 &67.6 &85.9 &45.8 & 62.7 \\
3DETR (our baseline) & 44.98	&82.84	&86.7	&89.66	&63.37	&46.57	&37.7	&43.38	&9.95	&59.84	&75	&60.19	&40.74	&60.71	&98.72	&68.23	&88.97	&46.34 & 61.33 \\
\textbf{3DETR + MSA} & \textbf{49.58}	&82.65	&\textbf{87.78}	&86.82	&\textbf{66.51}	&\textbf{47.48}	&34.78	&\textbf{43.73}	&\textbf{13.46}	&\textbf{61.42}	&\textbf{78.41} &57.21	&\textbf{49.29}	&\textbf{70.51}	&98.11	&65.80	&80.52	&\textbf{47.74} &\textbf{62.32} \\ 
\midrule
3DETR-m &52.71	&80.71	&89.75	&89.92	&66.04	&54.91	&38.14	&51.48	&13.52	&58.76	&75.08	&47.09	&56.6	&62.86	&94.34	&70.17	&85.96	&51.42 &64.88 \\
\textbf{3DETR-m + MSA} &51.08	&\textbf{82.66}	&\textbf{90.11}	&87.86	&65.85	&53.45	&\textbf{41.06}	&\textbf{55.00}	&\textbf{16.20}	&\textbf{60.33}	&\textbf{77.49}	&\textbf{55.66}	&53.96	&\textbf{68.38}	&\textbf{98.28}	&\textbf{73.31}	&\textbf{93.16}	&51.25 &\textbf{65.67} \\
\bottomrule
\end{tabular}%
}
\end{table*}

\begin{figure*}[hbt]
    \centering
    \subcaptionbox{Ground truth annotation%
    \label{fig:meshlab_output1:GT1}}[15em]%
  {%
    \includegraphics[width=15em]{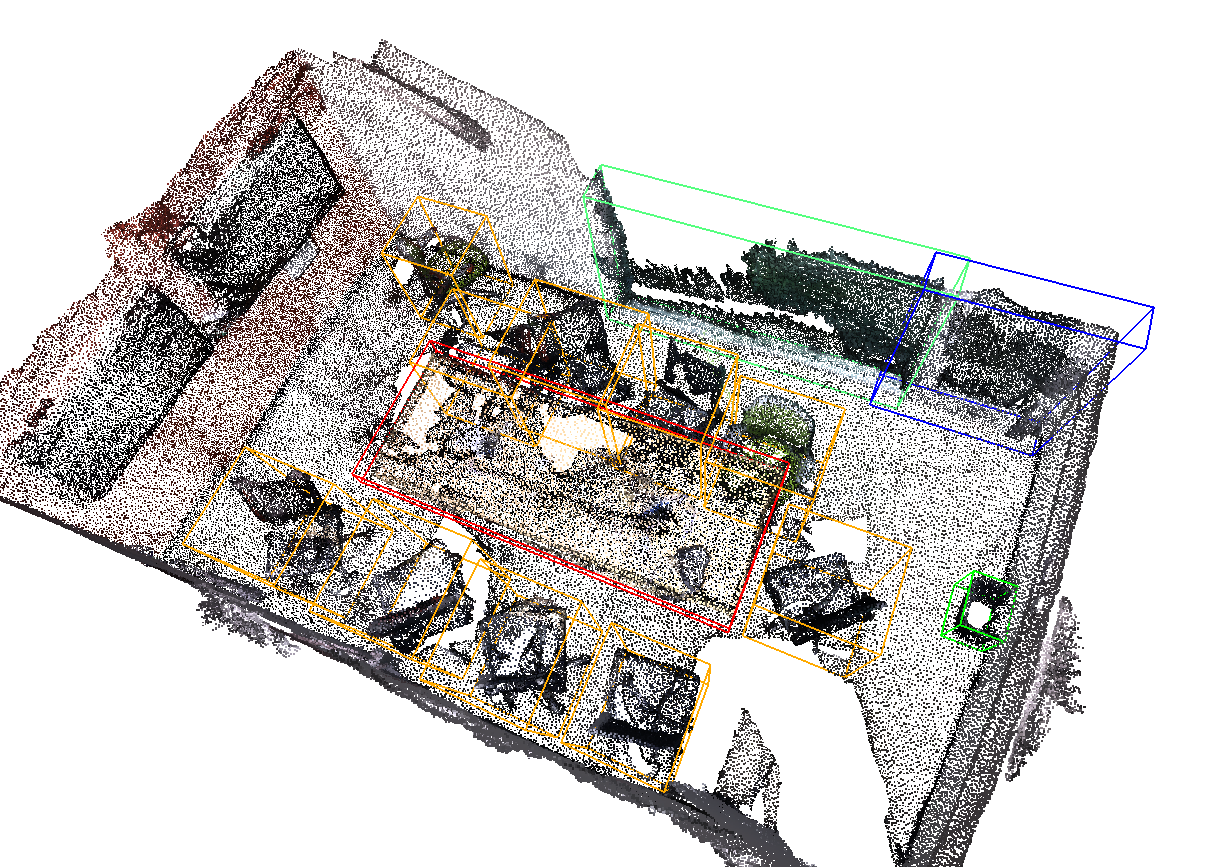}%
    }
    \hspace{1em}
    \subcaptionbox{3DETR%
    \label{fig:meshlab_output1:3DETR1}}[15em]%
  {%
    \includegraphics[width=15em]{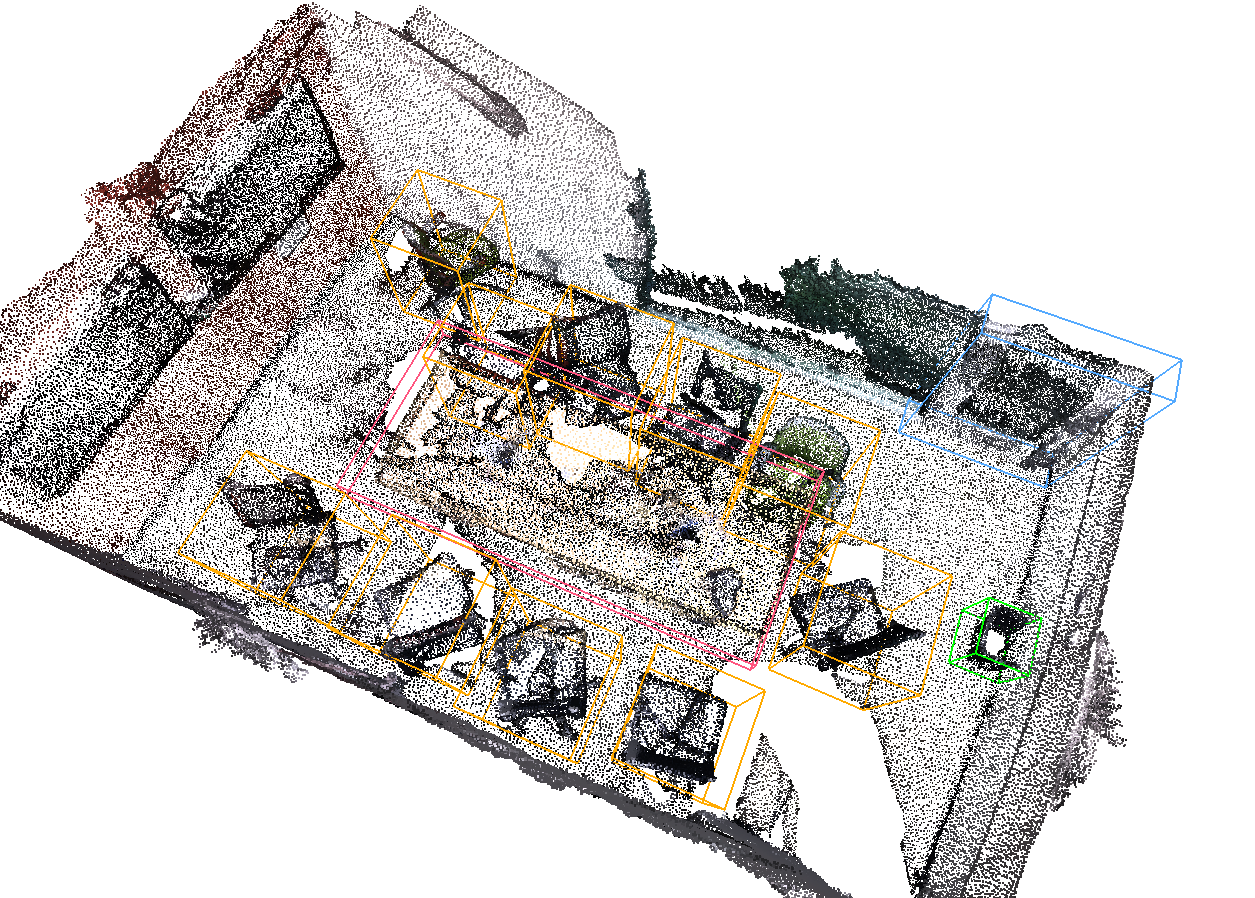}%
    }
    \hspace{1em}
    \subcaptionbox{3DETR $+$ MSA%
    \label{fig:meshlab_output1:MSA1}}[15em]%
  {%
    \includegraphics[width=15em]{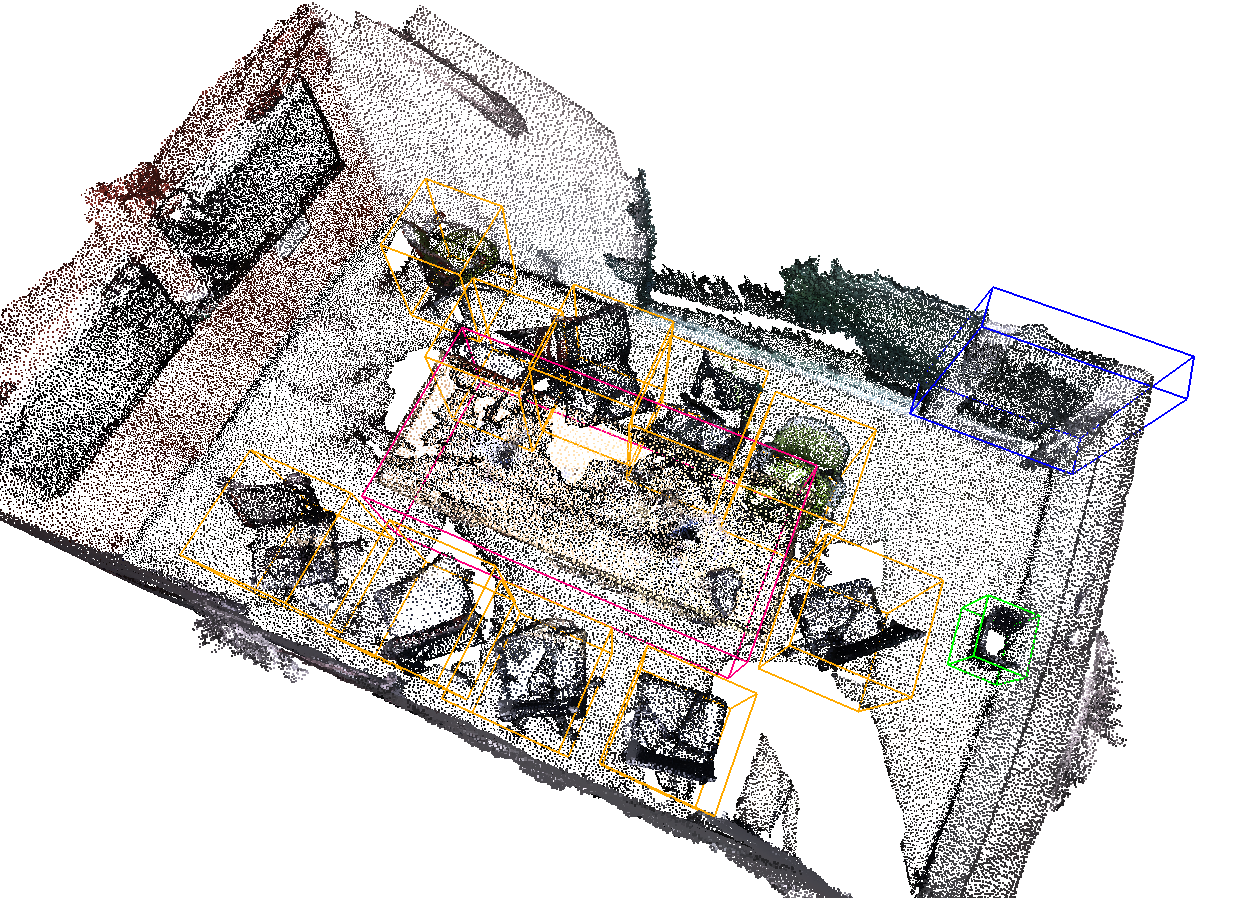}%
    }
    \caption{A sample output, where 
    a door is misclassified as a bookshelf and a window is not detected in~\subref{fig:meshlab_output1:3DETR1},
    while the door is properly classified in~\subref{fig:meshlab_output1:MSA1}.}
    \label{fig:meshlab_output1}
\end{figure*}


\begin{figure*}[hbt]
    \centering
    \subcaptionbox{Ground truth annotation%
    \label{fig:meshlab_output2:GT1}}[15em]%
  {%
    \includegraphics[width=15em, trim= 0px 25ex 0px 0px]{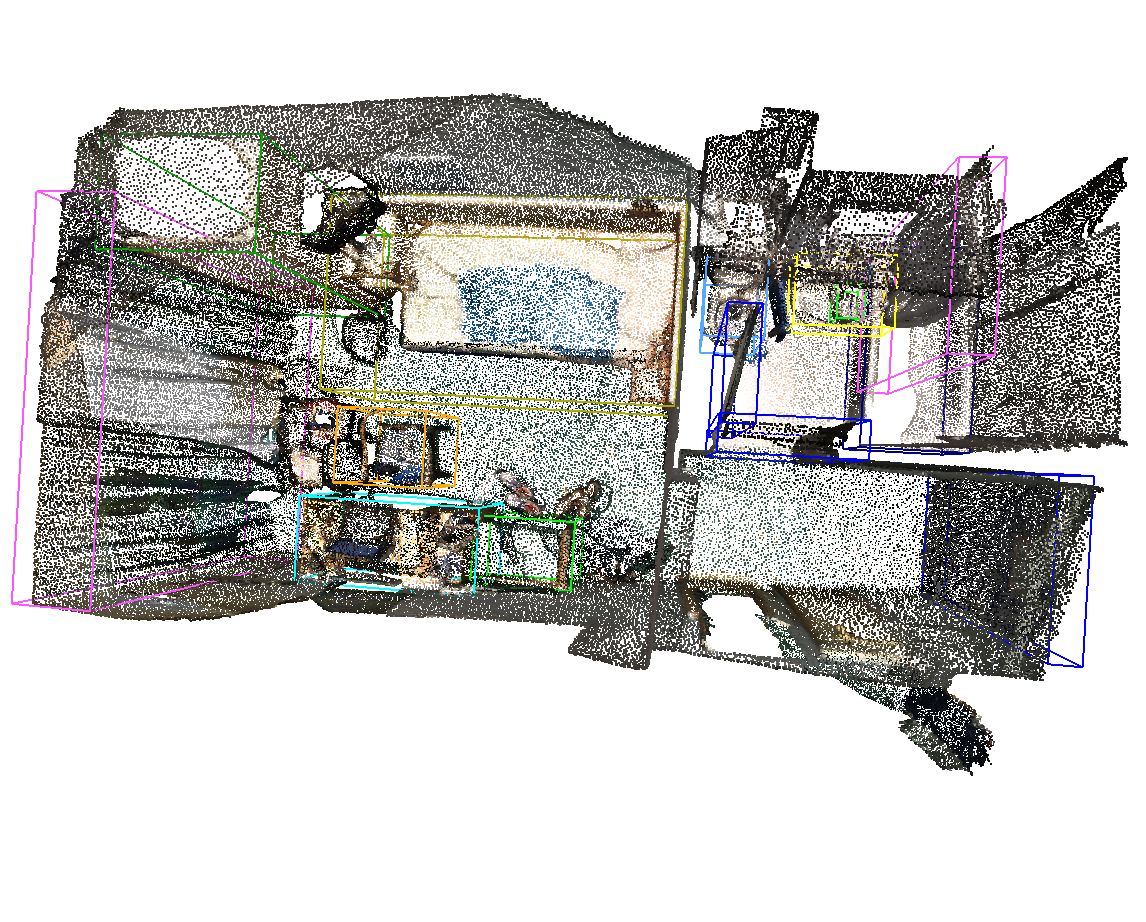}%
    }
    \hspace{1em}
    \subcaptionbox{3DETR%
    \label{fig:meshlab_output2:3DETR1}}[15em]%
  {%
    \includegraphics[width=15em, trim= 0px 25ex 0px 0px]{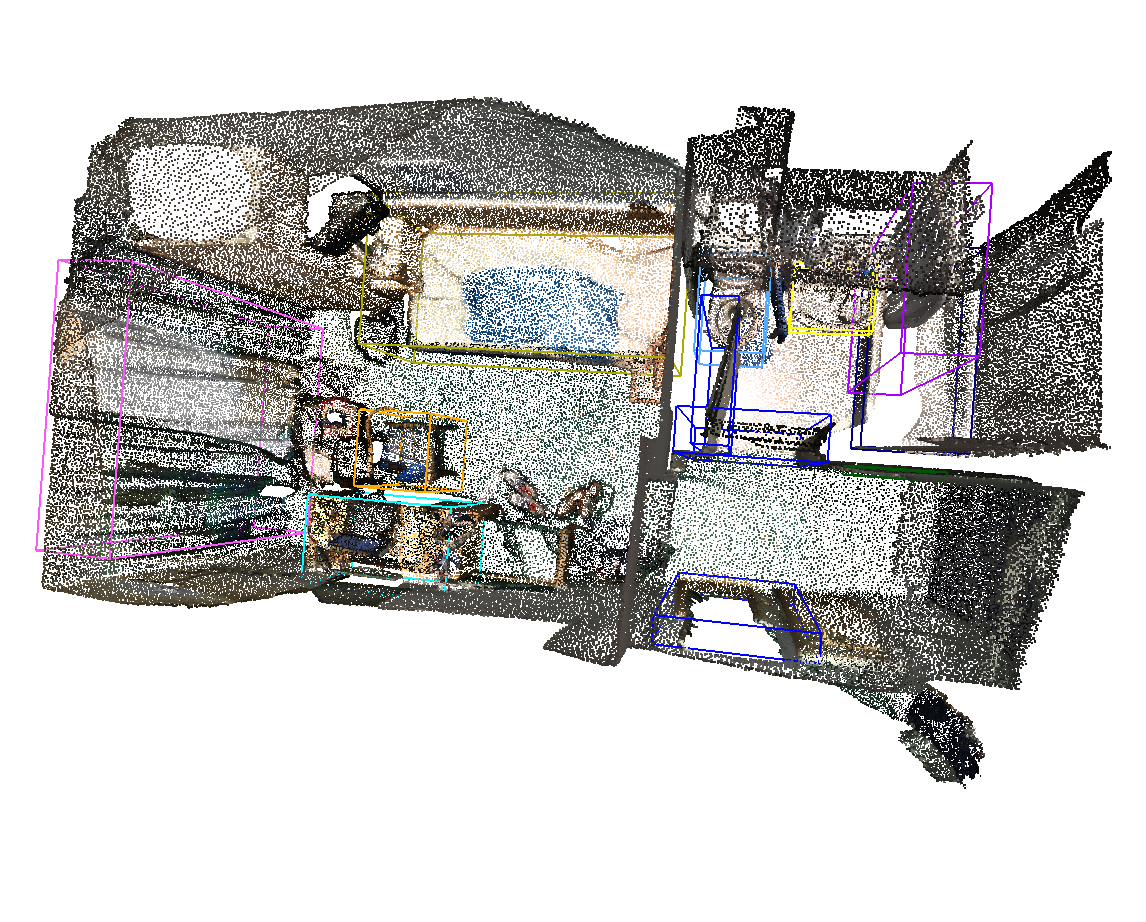}%
    }
    \hspace{1em}
    \subcaptionbox{3DETR $+$ MSA%
    \label{fig:meshlab_output2:MSA1}}[15em]%
  {%
    \includegraphics[width=15em, trim= 0px 25ex 0px 0px]{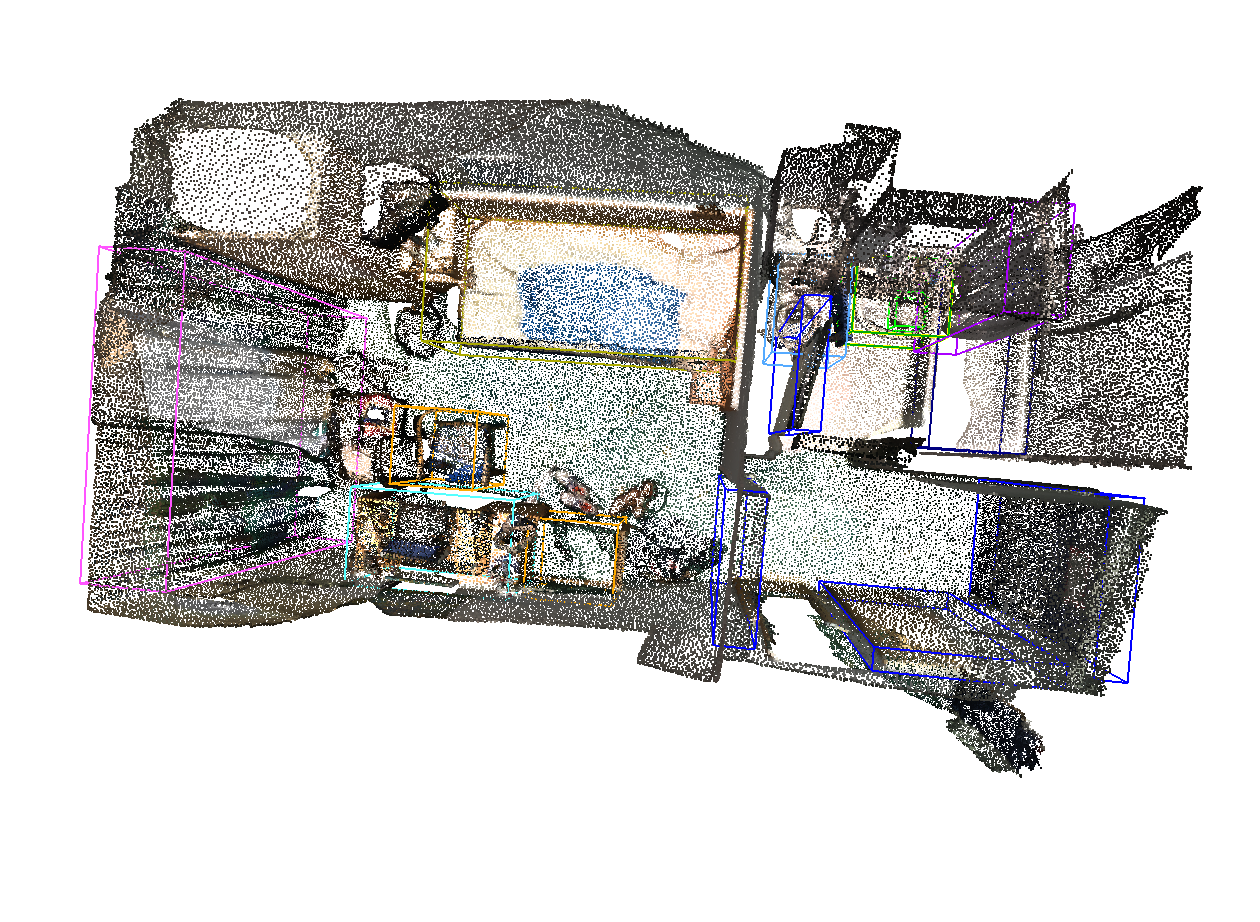}%
    }
    \caption{A sample output, where 
    a garbage bin and one of the doors are not detected in~\subref{fig:meshlab_output2:3DETR1},
    while the garbage bin is properly classified in~\subref{fig:meshlab_output2:MSA1}.}
    \label{fig:meshlab_output2}
\end{figure*}

\section{Experimental Validation}
\label{sec:experiments}
The proposed network is evaluated on the ScanNetv2 dataset~\cite{ScanComplete_2017}, which comprises highly detailed IoU listed in \Cref{tab:percat2}, and 3D reconstructions of 1,513 indoor scenes across 18 object categories listed in \Cref{tab:box_colors}. In addition to the spatial coordinates $(x,y,z)$, users can leverage color and height information as supplementary input features. The dataset provides comprehensive annotations, including 3D bounding boxes and per-point semantic instance labels.

\begin{table}[hbt]
    \centering
    \caption{The color map of object categories visualized in bounding boxes.}
\label{tab:box_colors}
    \begin{tabular}{|@{}l@{}c|l@{}c|l@{}c@{}|}
    \toprule
     \textbf{Object} & \textbf{Box color} &  \textbf{Object} & \textbf{Box color}  &  \textbf{Object} & \textbf{Box color} \\ 
     \midrule
    Cabinet  & \color{Green}{Green} &  Counter  & \color{Salmon}{Salmon} & Window  & \color{SeaGreen}{Sea Green} \\ 
    \midrule
    Bed  &  \color{LimeGreen}{Lime Green} & Desk  & \color{SkyBlue}{Sky Blue} & Sink  & \color{Yellow}{Yellow}  \\
    \midrule
    Chair  & \color{Orange}{Orange} &  Curtain  &  \color{VioletRed}{Violet Red} & Bookshelf  & \color{Periwinkle}{Periwinkle}  \\
    \midrule
    Sofa  & \color{Lavender}{Lavender} & Refrig.  &  \color{Bittersweet}{Bitter Sweet}  &  Bathtube  & \color{BlueViolet}{Blue Violet} \\
    \midrule
    Table  & \color{Rhodamine}{Rhodamine} &  Shower Curt. &  \color{Purple}{Purple} &  Picture  & \color{OliveGreen}{Olive Green} \\
    \midrule
    Door  & \color{Blue}{Blue} & Toilet  & \color{CadetBlue}{Cadet Blue}  &  Garb.bin  & \color{YellowGreen}{Yellow Green} \\
    \bottomrule
    \end{tabular}%
\end{table}

Following standard practice, evaluation uses the mean average precision metric (mAP) at intersection over union (IoU) thresholds of 0.25 and 0.5, without considering oriented bounding boxes. The data set is divided into 1,201 training scenes and 312 validation scenes. After training, evaluation is performed on the same validation set, with data points randomly shuffled during testing.

A total of 2,048 original points are fed into the encoder, while 4,096 points are input to the first set abstraction (SA1) layer of the PointNet++ backbone to obtain higher-resolution features. In terms of computational complexity, the proposed Multi-Scale Attention (MSA) module introduces minimal overhead, as it is applied only within the first transformer layer. However, due to the increased number of points processed, both training and inference times are slightly longer than those of the original model.

A per-category evaluation of the proposed architecture applied to the ScanNetV2 dataset, as measured by AP@25 IoU, is reported in \Cref{tab:percat2}, where bold results indicate improved performance relative to the baseline.
With the proposed network, improved detection performance is observed, particularly for smaller objects. For instance, compared to the paper results, detection accuracy increases by $+3.51$ for the picture category and by $+1.4$ for the garbage bin category. Enhanced performance is also noticeable for semantically related objects, such as table and chair, as well as desk, cabinet, and bookshelf. These object categories typically co-occur and are positioned near each other within scenes. This improvement suggests that the proposed Multi-Head Attention (MHA) mechanism more effectively captures local geometric cues without sacrificing the global scene context.

As a qualitative assessment, two sample outputs are presented in \Cref{fig:meshlab_output1,fig:meshlab_output2}, where the proposed 3DETR + MSA architecture demonstrates superior performance over the baseline 3DETR model by successfully detecting and correctly classifying objects that are otherwise missed or misclassified. It is important to note, however, that this improvement is accompanied by a slight increase in the number of false positives.

According to this, it appears to have more false positive examples, as it does not omit most objects, unlike garbage bins, as shown in Figure \ref{fig:meshlab_outputs}, whereas the 3DETR base model misses those object categories. Although the proposed model seems to have more false positive examples, it does not omit most objects like the base.

A quantitative evaluation is presented in \Cref{tab:mAP-results_sota}, reporting the Mean Average Precision (mAP) performance of several popular models trained on the ScanNetv2 dataset. The proposed 3DETR + MSA network achieves improvements of 62.32\% (+0.99) and 43.51\% (+4.78) in mAP@50 compared to the baseline model.

\begin{table}[htb]
    \centering
    \caption{Mean Average Precision (mAP) performance of popular models trained on the ScanNetv2 dataset.}
\label{tab:mAP-results_sota}
    \begin{tabular}{|@{}l|c|c|c@{}|}
    \toprule
     \textbf{Architecture} & \textbf{Backbone} & \textbf{mAP@25} & \textbf{mAP@50}  \\
    \midrule
    VoteNet & PN++ & 58.6 & 33.5  \\
    MLCVNet & PN++ & 64.5 & 41.4 \\
    Group-Free (L6, O256)  & PN++w2 $\times$ & 67.3 (66.2) & 48.9 (48.4) \\ 
    \midrule
    3DETR (paper) & SA1 & 62.7 & 37.5 \\
    3DETR (our baseline) & SA1 & 61.33 & 38.73 \\
    \textbf{3DETR + MSA} & SA1 & \textbf{62.32 (+0.99)} & \textbf{43.51 (+4.78)} \\ 
    \bottomrule
    \end{tabular}
\end{table}

\footnote{PN++ stands for PointNet++. PN++w2 refers to the backbone with a width of 2 in MLP networks. SA1 is the first layer of set abstraction in PointNet++.}

Applying the 3DETR-m model variant with MSA does not appear to improve performance, as summarized in \Cref{tab:mAP-results_sota2}. This outcome can be attributed to the interim downsampling stage that follows the encoder, where half of the original points are removed through masking. Consequently, this hampers the generation of higher-resolution feature maps. These results suggest that adapting the upsampling strategy is necessary when integrating MSA with 3DETR-m.

\begin{table}[hbt]
    \centering
    \caption{Mean Average Precision (mAP) performance of popular models trained on the ScanNetv2 dataset with the 3DETR-m model variant.}
\label{tab:mAP-results_sota2}
\vspace{0.1cm}
    \begin{tabular}{|@{}l|c|c|c@{}|}
    \toprule
     \textbf{Architecture} & \textbf{Backbone} & \textbf{mAP@25} & \textbf{mAP@50}  \\
         \midrule
    VoteNet & PN++ & 58.6 & 33.5  \\
    MLCVNet & PN++ & 64.5 & 41.4 \\
    Group-Free (L6, O256)  & PN++w2 $\times$ & 67.3 (66.2) & 48.9 (48.4) \\ 
    \midrule
    3DETR-m (paper) & SA1 & 65.0 & 47.0 \\
    3DETR-m (our baseline) & SA1 & 64.88 &46.58 \\
    \textbf{3DETR-m + MSA} & SA1 & \textbf{65.67 (+0.79)} & 45.35 \\ 
    \bottomrule
    \end{tabular}
\end{table}


\section{Conclusion}
\label{sec:conclusion}

In this work, we proposed a novel Multi-Scale Attention (MSA) mechanism for 3D object detection within point clouds, building upon the 3DETR architecture. By introducing an upsampling-based multi-head attention strategy, our method effectively captures finer local details while maintaining global context, leading to improved detection of smaller and semantically related objects. Experiments on the ScanNetv2 benchmark demonstrate that the proposed 3DETR+MSA model outperforms the baseline, particularly in the mAP@25 and mAP@50 metrics. However, when applied to the lightweight 3DETR-m variant, the current upsampling strategy proved less effective, indicating that further adaptation is necessary. Future work will explore advanced upsampling techniques and dynamic feature aggregation strategies to further enhance detection performance across varying model architectures.

\section*{Acknowledgments}
 This research was supported by the UAE University SURE PLUS-2025 Research.


%
%

\FloatBarrier
\printbibliography

\vfill

\end{document}